\documentclass[11pt]{article}

\usepackage[utf8]{inputenc}
\usepackage[margin=1in]{geometry}
\usepackage{amsmath,amssymb}
\usepackage{booktabs}
\usepackage{tabularx}
\usepackage{graphicx}
\usepackage{xcolor}
\usepackage{tikz}
\usetikzlibrary{shapes.geometric,arrows.meta,positioning}
\usepackage[round,authoryear]{natbib}
\usepackage{microtype}
\usepackage[hidelinks]{hyperref}
\usepackage{caption}
\newcommand{\thetaz}{\theta_0}
\newcolumntype{Y}{>{\raggedright\arraybackslash}X}

\title{Surprise Alone Learns Lies:\\Evidence-Gated Continual Learning for LLMs from Streaming Claims}

\author{
  Abdelkader Metales\\
  caaise GmbH\\
  \texttt{metales@caaise.com}
  \and
  Miriam Grabenheinrich\\
  caaise GmbH\\
  \texttt{grabenheinrich@caaise.com}
}

\date{}

\begin{document}
\maketitle

\begin{abstract}
Continual learning methods for large language models increasingly decide \emph{what} to learn by surprise, keeping the inputs the model finds least predictable. This rule is unsafe when the stream contains lies. A planted falsehood about something the model knows has low probability under the model's belief, so it is highly surprising to the model by construction, and a surprise-driven learner absorbs nearly every such lie. A learner that selects claims at random with the same training budget absorbs a similar share of lies. We study the problem with LoRA-adapted students (Qwen2.5-1.5B, Qwen2.5-3B and Llama-3.2-3B) that learn from a stream of true and false claims about invented people and real CounterFact facts.

We report four findings. (1)~A verifier key, a judge that must approve each claim, removes lies to the extent that its verifier is accurate. Real LLM judges are cautious and reject unsourced truths, and even source-backed updates of facts they believe otherwise. (2)~Calibrated regression tests on unrelated knowledge slow learning down without filtering out lies. (3)~The original model's own prior flags claims that contradict what it knows, without a judge. It separates lies about known facts from true new facts (AUC 0.78--0.83 across the three models) but not from legitimate updates of the same kind of fact (AUC 0.53), and the model's confidence in its own answer does as well. It saves 43\% of judge calls, but an attacker with access to the model can choose lies that pass it. (4)~If trusted sources exist that an attacker cannot plant, routing contradicted claims to evidence recovers legitimate updates, no lie enters through this route, and the combined system has the lowest poison rate among learned claims; when an attacker plants matching sources for 30\% of the lies, the route admits them and the system ends up worse than the judge alone. We identify this vulnerability as a central finding. The acceptance signals answer different questions, and none of them decides on its own whether a claim is true.
\end{abstract}

% ======================================================================
\section{Introduction}

Many deployed language models keep their base parameters fixed after training, even as applications increasingly need to incorporate new information. A model that could update itself from claims made by its users could stay current without retraining. An assistant for a university or a public administration, for example, must keep up with changing regulations and responsibilities without being retrained for each change. Catastrophic forgetting is one obstacle. Trust is the other, because anyone who can talk to such a model can try to teach it falsehoods. A 2016 chatbot that learned from users was corrupted within a day, and recent work shows that ordinary up/down feedback is enough to inject false facts into a preference-tuned model \citep{hilel2025hypnosis}.

Recent continual-learning methods for LLMs choose what to keep by \textbf{surprise}, the negative log-likelihood of an input \citep{hazard2025sure}. Surprise is a good proxy for novelty. It also makes a lie about a known fact stand out, because a model that holds the true belief gives the false claim a low probability. We asked what a surprise-driven learner absorbs when part of its stream is poison, and which additional checks restore safety without stopping learning.

We treat each update as a decision that must pass up to three checks, which we call keys, loosely inspired by how the brain gates memory consolidation: (i)~\emph{surprise}, is this new to me; (ii)~a \emph{verifier key}, does an external verifier approve it; and (iii)~a \emph{statistical test}, did learning it damage old knowledge, with commit or rollback of a LoRA adapter. We also use a fourth key that needs no judge, (iv)~a \emph{prior-conflict key} that holds back claims the original model confidently contradicts. The keys ask different questions. Surprise asks whether a claim is new, the prior whether it contradicts what the model believes, the verifier what the evidence says, and the regression test whether an update damaged unrelated knowledge. In our experiments, no key could stand in for another.

We ask four research questions, each answered in the sections shown:
\begin{enumerate}
  \item[\textbf{RQ1}] Does surprise-driven continual learning preferentially absorb false claims? (Section~\ref{sec:surprise})
  \item[\textbf{RQ2}] How do a verifier key, a regression test and the prior-conflict key affect poisoning and the learning of true claims, and what does each of them detect? (Sections~\ref{sec:judge}--\ref{sec:gemini} and~\ref{sec:detects})
  \item[\textbf{RQ3}] Can an evidence route recover the legitimate updates that the prior key blocks, without readmitting lies? (Section~\ref{sec:updates})
  \item[\textbf{RQ4}] Do these findings transfer across model size, model family and real judges? (Sections~\ref{sec:models}--\ref{sec:realjudges})
\end{enumerate}

This is a controlled study of factual claim poisoning in three models from two families. It does not offer a complete solution to continual-learning security. Our contributions are:
\begin{enumerate}
  \item A controlled benchmark of continual learning from a poisoned claim stream, with true facts the model cannot know (invented people), plausible known-fact lies (CounterFact counterfactuals) and lies about the unknown, measured with top-1 metrics whose chance level is known.
  \item Evidence that selecting by surprise does not filter poison and learns lies (84--93\% of all streamed lies asserted after learning: 99--100\% of known-fact lies, 68--86\% of lies about invented people) and that a verifier key removes them, which leaves the system as reliable as its judge and no more.
  \item A negative result: in our setting, a calibrated regression test on unrelated old knowledge throttles learning but does not filter poison, because a lie barely affects facts other than the one it targets.
  \item An analysis of a judge-free prior-conflict key. It cuts absorbed known-fact lies 2.5--5$\times$ with simulated judges and 1.5--2.5$\times$ with real ones and saves 43\% of judge calls. It detects contradiction rather than falsehood (lies vs.\ legitimate updates: AUC 0.53), does no better than the model's answer confidence, and is bypassed by an attacker who screens lies with the model. An evidence route recovers the legitimate updates it blocks, under a trusted-source assumption, and fails when sources can be planted.
\end{enumerate}

% ======================================================================
\section{Related work}

\paragraph{Continual learning for LLMs.} SuRe \citep{hazard2025sure} ranks inputs by surprise (high NLL) for replay and consolidates fast and slow LoRA adapters with an exponential moving average; its authors note that fully online, task-free learning remains open. STABLE \citep{hoy2025stable} gates sequential LoRA edits with a stability budget (exact-match drop, bits increase or KL divergence) and rejects or clips edits that exceed it; it names streaming inputs as future work. SEAL \citep{zweiger2025seal} lets a model write its own training data and keeps self-edits that improve a task reward. MEGa \citep{pan2025mega} stores each memory in its own gated LoRA adapter, inspired by complementary learning systems \citep{mcclelland1995cls}. K-Merge \citep{shenaj2026kmerge} merges incoming adapters under a storage budget. PS-LoRA \citep{zhou2025pslora} limits interference between sequential adapters. All of these decide on novelty, stability or task reward; none asks whether an incoming fact is \emph{true}. Our surprise-only and stability-only baselines are simplified versions of SuRe and STABLE.

\paragraph{Poisoning through user feedback.} LLM Hypnosis \citep{hilel2025hypnosis} shows that a single unprivileged user can inject persistent false facts through prompts and preference feedback. \mbox{CAPTURE} \citep{hossain2026capture} separates genuine preference drift from poisoning in agent memory, but leaves weight-level learning untouched. OWASP lists data and model poisoning as LLM04:2025 \citep{owasp2025llm}.

\paragraph{Retrieval instead of learning.} Retrieval-augmented generation \citep{lewis2020rag} keeps new facts outside the weights, so a poisoned fact never becomes part of the model. It does not remove the trust problem, because a poisoned document store misleads the model at inference time just as a poisoned stream does during learning. We study the parametric case, where an accepted claim is written into the weights and persists in every later answer, which makes the acceptance decision more consequential. The evidence route of Section~\ref{sec:updates} sits between the two: it uses a retrieved source to decide what may be written into the weights.

\paragraph{Verification and knowledge editing.} LLM-as-a-judge \citep{zheng2023judge} is the standard way to score model outputs; we use it as a gate on learning. CounterFact \citep{meng2022rome} provides real facts paired with counterfactual rewrites and is the standard benchmark for knowledge editing; we reuse its counterfactuals as plausible lies. \citet{li2025keti} identify, after the fact, whether a model was edited maliciously (for example with misinformation) or benignly (fact updates); we ask the same question before a claim is learned, and Section~\ref{sec:detects} shows that the model's prior alone cannot answer it.

\paragraph{Our contribution.} Judges that override a provided reference when it contradicts their own knowledge have been documented at evaluation time \citep{lee2026judging}, and claim-level provenance with hadith-style chains of narrators has been proposed for multi-agent knowledge systems \citep{raja2026isnad}. We study the \emph{acceptance} decision when claims are written into the weights of a continually learning model and the stream is poisoned by a fixed attacker; we additionally evaluate one adaptive attacker that screens its lies against the prior key. Our contributions are empirical: the vulnerability of surprise-driven selection, and a measurement of what each acceptance signal detects, including how it fails.

% ======================================================================
\section{Method}
\label{sec:method}

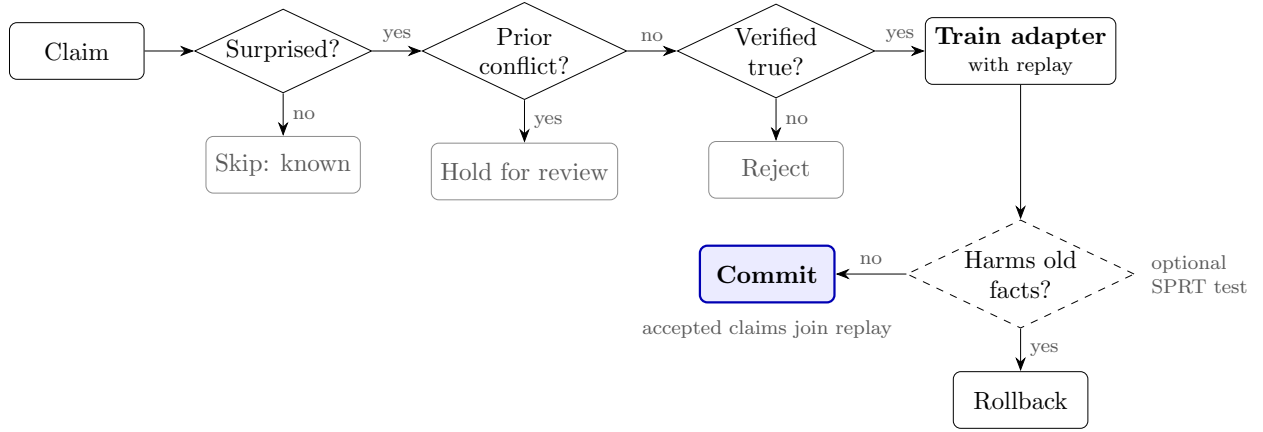
\begin{figure}[htbp]
\centering
\resizebox{\linewidth}{!}{%
\begin{tikzpicture}[
  font=\small,
  node distance=6mm and 7mm,
  box/.style={draw, rounded corners=3pt, minimum height=8mm, minimum width=19mm, align=center},
  exit/.style={box, text=black!65, draw=black!45},
  key/.style={draw, diamond, aspect=2.1, inner sep=1pt, align=center, minimum width=25mm},
  main/.style={box, draw=blue!70!black, fill=blue!8, line width=0.9pt},
  lab/.style={font=\scriptsize, text=black!65},
  >={Stealth[length=2mm]}
]
\node[box]                           (claim)  {Claim};
\node[key, right=of claim]           (surp)   {Surprised?};
\node[key, right=of surp]            (prior)  {Prior\\conflict?};
\node[key, right=of prior]           (truth)  {Verified\\true?};
\node[box, right=of truth]           (train)  {\textbf{Train adapter}\\[-1pt]{\scriptsize with replay}};
\node[exit, below=of surp]           (skip)   {Skip: known};
\node[exit, below=of prior]          (hold)   {Hold for review};
\node[exit, below=of truth]          (reject) {Reject};
\node[key, dashed, below=19mm of train] (test) {Harms old\\facts?};
\node[main, left=10mm of test]       (commit) {\textbf{Commit}};
\node[box, below=of test]            (rollback) {Rollback};
\node[lab, below=1mm of commit]      {accepted claims join replay};
\node[lab, right=1mm of test.east, align=left] {optional\\SPRT test};

\draw[->] (claim) -- (surp);
\draw[->] (surp)  -- node[lab, above] {yes} (prior);
\draw[->] (prior) -- node[lab, above] {no}  (truth);
\draw[->] (truth) -- node[lab, above] {yes} (train);
\draw[->] (surp)  -- node[lab, right] {no}  (skip);
\draw[->] (prior) -- node[lab, right] {yes} (hold);
\draw[->] (truth) -- node[lab, right] {no}  (reject);
\draw[->] (train) -- (test);
\draw[->] (test)  -- node[lab, above] {no}  (commit);
\draw[->] (test)  -- node[lab, right] {yes} (rollback);
\end{tikzpicture}}
\caption{The acceptance decision for one claim (\texttt{sura/runner.py}). A claim is learned only if it is new, uncontradicted and verified. Claims are processed in chunks of 16; each key can be switched off for the ablations.}
\label{fig:keys}
\end{figure}

A frozen base model with parameters $\thetaz$ learns only through a LoRA adapter. Claims $c = (x, y)$, a prompt $x$ and an object $y$, arrive in chunks. For each chunk, every claim must pass the active keys; the accepted claims, together with replayed claims accepted earlier, train a candidate adapter, which is then committed or rolled back (Figure~\ref{fig:keys}). Each key measures a different property, and none of them measures truth directly: surprise measures novelty, the prior-conflict key contradiction with what the original model knows, the verifier key a judge's verdict, the statistical test collateral damage to unrelated knowledge, and the evidence route agreement with a source.

\paragraph{Surprise.} The mean negative log-likelihood per target token under the current model,
\begin{equation}
s_\theta(c) = -\frac{1}{|y|} \sum_{t=1}^{|y|} \log p_\theta(y_t \mid x, y_{<t}).
\end{equation}
A claim is a learning candidate when $s_\theta(c)$ exceeds $\tau = 1.0$ nats/token. Claims the model already predicts are skipped.

\paragraph{Prior-conflict key.} We ask the \emph{original} model $\thetaz$ for its own greedy answer $\hat{y}$ to the prompt and measure how much less likely the claim is than that answer:
\begin{equation}
g(c) = s_{\thetaz}(c) - s_{\thetaz}(\hat{y}), \qquad \text{hold back if } g(c) \ge \tau_g = 6.
\end{equation}
A claim with a large gap contradicts a belief the model held before any user taught it anything, so it is held back for review instead of learned. We first computed the gap under the current adapter. In a pilot this held back 46\% of true claims, because during learning the adapter becomes overconfident in recently learned patterns; switching to $\thetaz$ removed the problem. The gap needs no judge and is computed once per claim.

\paragraph{Verifier key.} A verifier sees the claim, plus a trusted source sentence when one exists, and answers true or false. Each distinct claim is judged once.

\paragraph{Statistical test (optional).} After training a candidate adapter, we score the regression suite in batches of $b = 25$ items. For item $i$, $d_i$ is the loss of preference margin (nats/token) for the true object over its counterfactual, relative to the original base $\thetaz$. We compare $H_0\colon \mathrm{mean}(d) = \mu_0$, the drift of clean learning calibrated in Section~\ref{sec:regression}, against $H_1\colon \mathrm{mean}(d) = \mu_0 + \delta$, using a Gaussian log-likelihood ratio after $n$ scored items with Wald's boundaries \citep{wald1945sprt}:
\begin{equation}
\Lambda_n = \frac{\delta}{\hat\sigma_n^2}\left(\sum_{i=1}^{n} d_i - n\left(\mu_0 + \frac{\delta}{2}\right)\right), \qquad A = \ln\frac{1-\beta}{\alpha}, \qquad B = \ln\frac{\beta}{1-\alpha}.
\end{equation}
After each batch, $\Lambda_n \ge A$ ($2.77$ for $\alpha = 0.05$, $\beta = 0.2$) declares harm and $\Lambda_n \le B$ ($-1.56$) declares no harm. If the 200-item suite runs out first, harm is declared when $\Lambda_n > 0$. Harm restores the adapter and optimizer state from their snapshot. The variance $\hat\sigma_n^2$ is re-estimated from all items scored so far, items are treated as independent, and the suite is finite. The procedure is therefore SPRT-style: Wald's boundaries set its thresholds, but we claim no formal control of the error rates $\alpha$ and $\beta$. An alternative paired one-sided $t$-test on the whole suite is implemented but not reported.

\paragraph{Training.} Cross-entropy on the target tokens of the accepted and replayed claims plus $\mathrm{KL}(\thetaz \,\|\, \theta)$ on anchor sentences of known facts, which keeps the adapter close to the base. The replay buffer keeps the 512 most surprising accepted claims.

% ======================================================================
\section{Experimental setup}
\label{sec:setup}

\paragraph{Student.} Qwen2.5-1.5B-Instruct \citep{qwen2024qwen25} (Qwen2.5-3B-Instruct and Llama-3.2-3B-Instruct \citep{llama2024llama3} for the replication in Section~\ref{sec:models}), frozen, in bfloat16, with one LoRA adapter \citep{hu2022lora} (rank 16, alpha 32, on all attention and MLP projections). Each accepted chunk trains for 8 AdamW steps (learning rate $2\times10^{-4}$, batch 16) on its claims plus 32 replayed accepted claims, with a KL term to the frozen base on 16 anchor sentences. Everything runs on one 16\,GB laptop GPU; a run over the stream takes 5--12 minutes.

\paragraph{Data.} Table~\ref{tab:data} lists the claim sets. The generator gives every one of the 120 people all 5 relations, a value drawn from that relation's pool of 12--18 values, and shuffles the 600 facts with the world seed. The first 400 become the true new facts; the remaining 200 form the pool for lies about invented people, so no lie concerns a fact that is also streamed as true. A poison rate $p$ adds $\mathrm{round}(p/(1-p) \times 400)$ lies, split evenly between the two kinds (171 lies at 30\%, 400 at 50\%). The poison rate is thus the share of lies among the new facts and lies in the stream; the 100 legitimate updates of Sections~\ref{sec:updates}--\ref{sec:realjudges} come on top. The same 400 new facts are used at every poison rate, so rates compare fairly. Sections~\ref{sec:updates}--\ref{sec:realjudges} add 100 legitimate updates, taken from known facts outside every other set. Each claim appears twice in a shuffled stream, read in chunks of 16. Retention is measured only on the 400 held-out facts, which no system ever sees. We use atomic factual claims on purpose, to isolate the acceptance decision from the confounds of long documents, multi-hop reasoning and task-specific objectives. We call the CounterFact claims known-fact lies: their subjects and relations are real, but the claims are counterfactual rewrites, not naturally occurring misinformation.

\begin{table}[htbp]
\centering\small
\caption{Claim sets of one world.}
\label{tab:data}
\begin{tabularx}{\linewidth}{@{}lYlY@{}}
\toprule
Set & Source & Size & Role \\
\midrule
New facts & 400 of the 600 facts generated for 120 invented people $\times$ 5 relations (birthplace, profession, language, instrument, sport) & 400 & True claims the model cannot know; should be learned \\
Lies about invented people & The other 200 generated facts, each given a wrong value from its relation's pool & up to 200 & Lies a source would contradict \\
Known-fact lies & CounterFact \citep{meng2022rome} counterfactuals on facts the student knows & up to 200 & Plausible lies about known facts \\
Known facts & CounterFact records the student ranks correctly (2,443 of 3,000 screened) & 200 / 300 / 400 & Regression suite / KL anchor / held-out retention set \\
\bottomrule
\end{tabularx}
\end{table}

\paragraph{Development and held-out data.} Seeds 0--2 (Sections~\ref{sec:surprise}--\ref{sec:gemini}) served development: the gate's null drift was calibrated on clean seed 0--2 runs, and the prior-key rule and its threshold, $g \ge 6$, were chosen on a 30-claim sample of the seed-0 world, which the prior sweep of Section~\ref{sec:prior} also uses. The seed-7 diagnostic (Figure~\ref{fig:tradeoff}, Table~\ref{tab:operating}) and every run on seeds 10--14 (Sections~\ref{sec:updates}--\ref{sec:realjudges}) were made after the threshold was fixed and never retuned it, also for the two further models; we treat them as post-development evaluations with the threshold fixed. Because the development sample overlaps the seed-0 world, the prior sweep of Section~\ref{sec:prior} is not an unbiased estimate of the threshold's performance; the cleaner evidence comes from seed 7 and seeds 10--14. The overlap analysis at the end of Section~\ref{sec:realjudges} is exploratory.

\paragraph{Systems.} Each system is the same loop with keys switched on or off (Table~\ref{tab:systems}).

\begin{table}[htbp]
\centering\small
\caption{Systems compared. Every system uses the same student, stream and training loop.}
\label{tab:systems}
\begin{tabularx}{\linewidth}{@{}lYY@{}}
\toprule
System & Learns from & Safety check \\
\midrule
Naive & every claim & none \\
Surprise-only (SuRe-lite) & surprising claims, replay, fast/slow EMA & none \\
Stability-only (STABLE-lite) & every claim & rollback if suite exact-match drops $>0.02$ \\
Verifier key & surprising and verified claims & none \\
Verifier key + test & surprising and verified claims & SPRT vs.\ original base, commit / rollback \\
Test only & surprising claims & SPRT (ablation: no verifier key) \\
Prior only & surprising claims without prior conflict & none (no judge) \\
Verifier key + prior & surprising, verified, no prior conflict & none \\
Verifier key + prior + evidence route & as above, plus held-back claims whose trusted source states the same value & none \\
\bottomrule
\end{tabularx}
\end{table}

\paragraph{Verifiers.} An oracle (ground truth), a simulated judge that is right with probability 0.6--0.9 and wrong consistently on the same claims, and Gemini 2.5 Flash \citep{gemini2025gemini25}, which sees a trusted source sentence for the invented people that have one. Two local models served through Ollama, phi4 (14B) \citep{abdin2024phi4} and Llama 3.2 (3B) \citep{llama2024llama3}, act as real judges in Section~\ref{sec:realjudges}. All real judges receive the same prompt, given in Appendix~\ref{app:prompt}.

\paragraph{Metrics.} For invented people, a fact counts as learned, or a lie as absorbed, when it is the model's top-1 choice among all 12--18 values of its relation (chance about 6\%). For real facts we compare the true object with the counterfactual (before learning: 1 for retention, 0 for known-fact lies). Lies absorbed is measured on the model after learning, over every distinct lie $\ell$ in the set $L$ of streamed lies, not only those selected for training. We report it separately for known-fact lies and lies about invented people; ``all lies'' is their count-weighted mean. Lies per 1,000 learned claims instead measures the learner's input: it counts claim occurrences accepted for training in committed chunks, excluding replay.
\begin{gather}
\text{absorbed} = \frac{|\{\ell \in L : \text{the model asserts } \ell \text{ after learning}\}|}{|L|},\\
\text{lies per 1{,}000} = 1000 \cdot \frac{\#\,\text{lie occurrences trained}}{\#\,\text{claim occurrences trained}}.
\end{gather}
Results are means over 3 seeds in Sections~\ref{sec:surprise}--\ref{sec:gemini} and over 5 held-out seeds in Sections~\ref{sec:updates}--\ref{sec:realjudges}. Because claims about the same invented person are not independent, uncertainty estimates that involve them resample whole people (person-clustered bootstrap).

% ======================================================================
\section{Results}

\subsection{Surprise alone learns lies}
\label{sec:surprise}

\begin{figure}[htbp]
\centering
\includegraphics[width=0.72\linewidth]{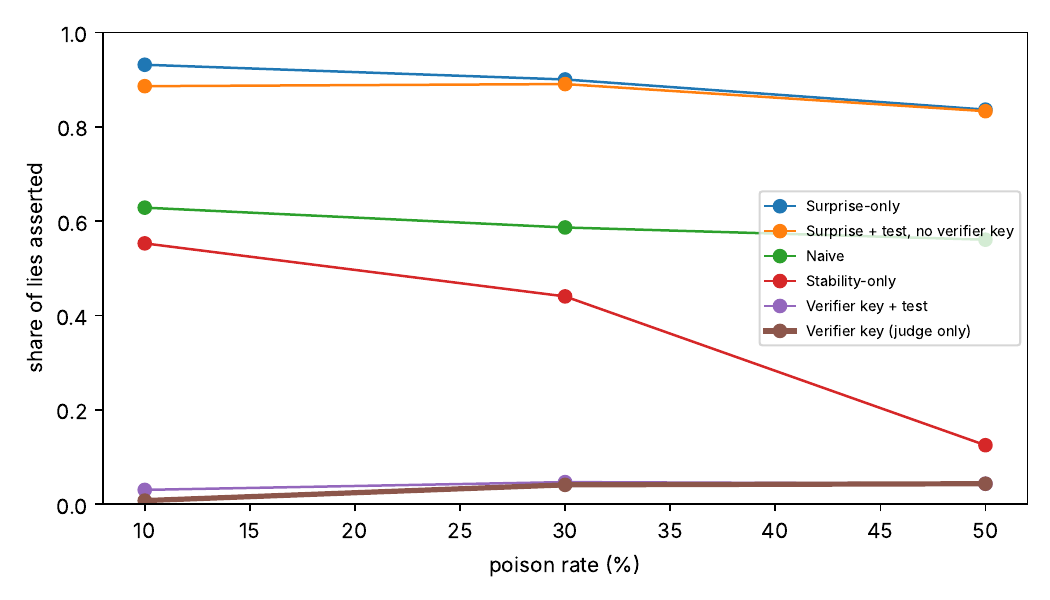}
\caption{Share of all streamed lies asserted after learning, by poison rate (oracle verifier; Qwen2.5-1.5B; 6 systems $\times$ 3 poison rates $\times$ 3 seeds; \texttt{results/main}).}
\label{fig:poison}
\end{figure}

With a perfect verifier, every system without a verifier key learned most of the lies it saw. The two surprise-driven systems learned the most, since lies are among the inputs they select. Known-fact lies fared worst. Surprise-only learning absorbed 99--100\% of them at every poison rate, against 68--86\% of lies about invented people and 84--93\% of all lies (the measure in Figure~\ref{fig:poison}).

Table~\ref{tab:main} compares the systems at 30\% poison. The verifier key learned at least as much as any other system and protected old knowledge best.

\begin{table}[htbp]
\centering\small
\caption{Learning and retention at 30\% poison (oracle verifier). Means over 3 seeds; seed standard deviations are at most 0.05, except for the verifier key with the uncalibrated test (0.18 at 30\%, 0.35 at 50\% poison), whose rollbacks made learning unstable.}
\label{tab:main}
\begin{tabular}{@{}lcc@{}}
\toprule
System (30\% poison) & New facts learned & Old facts kept \\
\midrule
Before learning & 0.06 & 1.00 \\
Naive & 0.19 & 0.92 \\
Surprise-only & 0.80 & 0.93 \\
Stability-only & 0.17 & 0.90 \\
Surprise + test, no verifier key & 0.74 & 0.92 \\
\textbf{Verifier key} & \textbf{0.86} & \textbf{0.98} \\
Verifier key + uncalibrated test & 0.49 & 0.98 \\
\bottomrule
\end{tabular}
\end{table}

The verifier key also kept more old knowledge (0.98 vs.\ 0.93), probably because rejecting lies about known facts avoids the edits that would overwrite them most directly. Naive learning, without replay, retains little of what it sees. The stability-only gate protects old facts but blocks most learning; at 50\% poison it rolled back 94 of 100 chunks.

\paragraph{Surprise versus random selection.} To separate the effect of selecting by surprise from the effect of training at all, we added a control that trains on a random subset of each chunk, of the same size as the surprising claims, with the same replay and averaging (held-out seeds 10--14, 30\% poison, 100 updates, sources for half the claims, judge accuracy 0.9 for the judged systems). Both learners took in lies at the base rate of the stream, 254 and 257 per 1,000 learned claims (paired difference $-3$, 95\% CI $-15$ to $+9$). Selecting by surprise neither raised nor lowered the share of lies in the training input. On known-fact lies it did worse than random selection. It absorbed 99.5\% of them against 88.9\% (paired difference $+11$ points, 95\% CI $+9$ to $+13$) while training on 86 fewer claims per run, because a lie about a known fact is always surprising. Against poison, a surprise filter is thus no better than random selection, and for lies about known facts it is worse.

\paragraph{Open-ended answers.} In the same runs we also let the model answer every prompt freely (greedy decoding) and counted a claim as asserted when the answer began with its value. Under this exact-match assertion rule, the free answers are directionally consistent with the ranking metrics. The surprise-only learner stated 98\% of the known-fact lies (ranking metric: 99.5\%), the verifier key 14\% (17\%) and the full system with the evidence route 4\% (8\%); before learning, the model stated none of them. True new facts were stated in 71--79\% of cases. Paraphrased questions were answered with the learned value less often (39--41\% for new facts, 3--37\% for known-fact lies), so facts learned by fine-tuning transfer weakly across wordings. Retention cannot be read this way: before learning, the model's free answers matched the CounterFact object for only 21\% of the facts it knew.

\subsection{Lies absorbed follow the judge's errors}
\label{sec:judge}

\begin{figure}[htbp]
\centering
\includegraphics[width=0.55\linewidth]{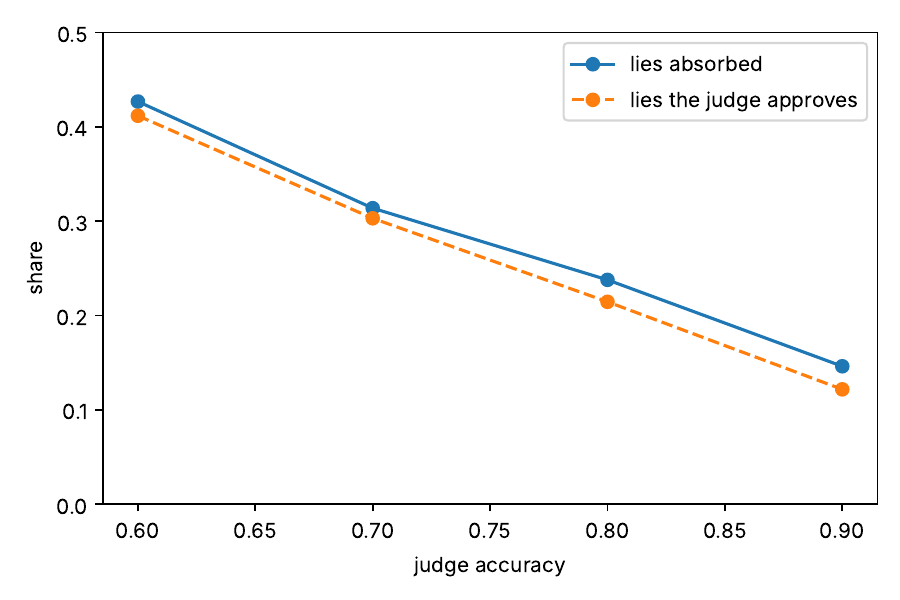}
\caption{Lies absorbed by the verifier key without test, against the share of lies the simulated judge approves (4 accuracies $\times$ 3 seeds; \texttt{results/judge\char95{}sweep}).}
\label{fig:judge}
\end{figure}

The simulated judge in this section is not meant as a realistic model of judge behaviour. It is a controlled sensitivity analysis that isolates the effect of verifier accuracy; Section~\ref{sec:realjudges} provides the realism check with real judges. Real verifiers make mistakes, so we replaced the oracle with a simulated judge that is right with probability 0.6 to 0.9. Its errors are consistent, so a lie that fools it once always fools it. Lies absorbed then followed the judge almost one for one (Figure~\ref{fig:judge}). The student learned nearly every lie the judge approved and very few others.

Learning of new facts also improved with judge quality (0.60 at accuracy 0.6, 0.80 at 0.9), because a weaker judge rejects more true claims. A verifier key shifts the problem to the verifier. The safety of continual learning becomes the verifier's reliability, measured on the claims an attacker chooses.

\subsection{Regression tests throttle, they do not filter}
\label{sec:regression}

The statistical test can detect that something went wrong, but not which claim caused it. We calibrated it on clean learning (no poison, test off). The per-chunk mean loss of preference margin on the suite ranged from $-0.72$ to $+0.78$ nats/token while retention stayed at 0.97--0.98, so we set the null drift to its 95th percentile (0.446) and the harmful effect to its standard deviation (0.203). With this calibration, rollbacks rise as the judge weakens, but the share of poison among the claims that are learned does not change (Table~\ref{tab:regression}).

\begin{table}[htbp]
\centering\small
\caption{The calibrated test with a simulated judge at 30\% poison (3 seeds; \texttt{results/judge\char95{}sweep}).}
\label{tab:regression}
\begin{tabular}{@{}cccccc@{}}
\toprule
& Rollbacks & \multicolumn{2}{c}{Poison share of learned claims} & \multicolumn{2}{c}{New facts learned} \\
\cmidrule(lr){3-4}\cmidrule(lr){5-6}
Judge accuracy & per run (with test) & with test & without test & with test & without test \\
\midrule
0.6 & 26.3 & 22.0\% & 22.2\% & 0.46 & 0.60 \\
0.7 & 38.3 & 14.7\% & 15.3\% & 0.40 & 0.67 \\
0.8 & 15.3 & 10.2\% & 9.8\% & 0.63 & 0.71 \\
0.9 & 2.0 & 5.3\% & 5.3\% & 0.79 & 0.80 \\
\bottomrule
\end{tabular}
\end{table}

A rolled-back chunk mixes accepted true claims with leaked lies, and the test discards both. It lowers lies absorbed only by learning less (at accuracy 0.7, lies fall from 0.31 to 0.20 and new facts from 0.67 to 0.40), which a smaller learning rate would also achieve. The earlier uncalibrated version, which tested the raw NLL rise with a tolerance of 0.05, was worse. It rolled back over half the chunks even with no poison and halved learning (Table~\ref{tab:main}).

We built the test as the main safety check, so this result was a disappointment, and the reason seems simple. A lie about one fact barely moves the model's answers on unrelated facts, and our suite contained only unrelated facts. In our setting, regression testing on unrelated knowledge did not filter targeted factual poisoning. A suite built from the neighbourhood of each incoming claim, or designed to probe contradictions, might behave differently; we did not test this.

\subsection{The model's own prior catches contradictions}
\label{sec:prior}

\begin{figure}[htbp]
\centering
\includegraphics[width=0.55\linewidth]{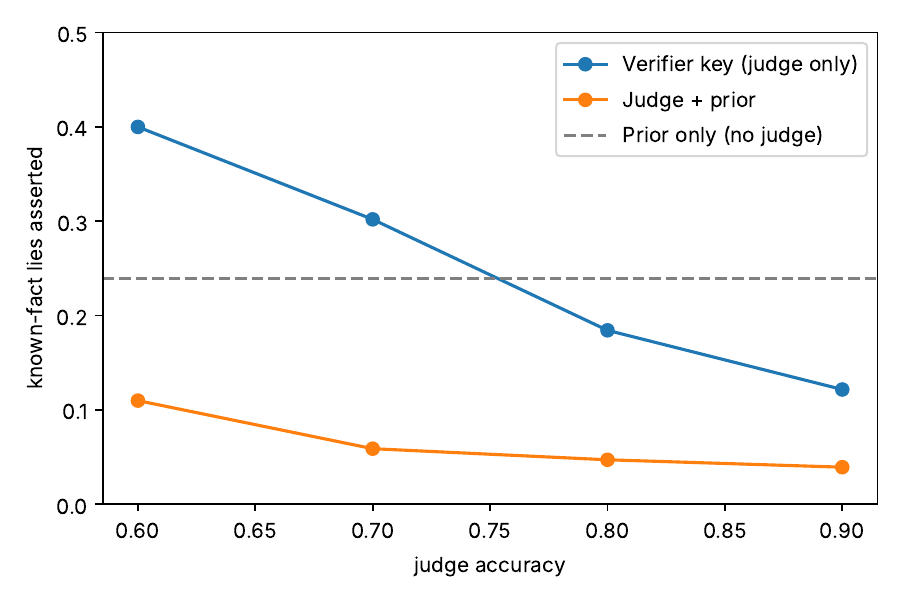}
\caption{Known-fact lies asserted after learning, with and without the prior key ($g \ge 6$), across judge accuracies (3 seeds; \texttt{results/prior\char95{}sweep}).}
\label{fig:prior}
\end{figure}

A lie about a known fact contradicts something the model already believes, and a true new fact does not. The prior-conflict key uses this difference without any judge. Added to the imperfect judge, it cut the known-fact lies the student ends up asserting by 3--5$\times$ at every accuracy, so a weak judge with the prior key did better than a good judge alone (Figure~\ref{fig:prior}). With no judge at all, the prior key held known-fact lies to 24\%, against 99--100\% for surprise-only learning.

The key has two costs. It held back 30\% of true new claims in the sweep, which lowered learning of new facts from 0.60--0.79 to 0.44--0.60 depending on the judge; in a deployed system these claims would go to review rather than be lost. Retention was unchanged or slightly better (0.97--0.98). The key also cannot see lies about invented people, which were absorbed about as often with it as without it (0.35 vs.\ 0.41 at judge accuracy 0.6, 0.15 vs.\ 0.16 at 0.9), since the model holds no belief for them to contradict.

To measure the trade-off independently of the sweep, we scored 200 claims of each type on a fresh world (seed 7) under the original model. The gap separates known-fact lies from true new facts with AUC 0.83 (95\% percentile bootstrap CI 0.79--0.87 from 300 resamples of single claims; resampling whole invented people with all their claims, 1,000 resamples, also gives 0.79--0.87), and lies about invented people no better than chance (AUC 0.49, CI 0.43--0.56). The operating point used in the sweep, $g \ge 6$, holds back 27\% of true facts and 76\% of known-fact lies on this fresh world, matching the 30\% cost measured in the sweep; with equal numbers of each, 74\% of the claims it holds back are lies. Measured as the share of lies among the claims actually learned, the prior key lowers poison from 152 to 111 per 1,000 learned claims at judge accuracy 0.7 and from 54 to 43 at 0.9; the remaining lies are mostly about invented people, which the key cannot see.

\begin{figure}[htbp]
\centering
\includegraphics[width=0.45\linewidth]{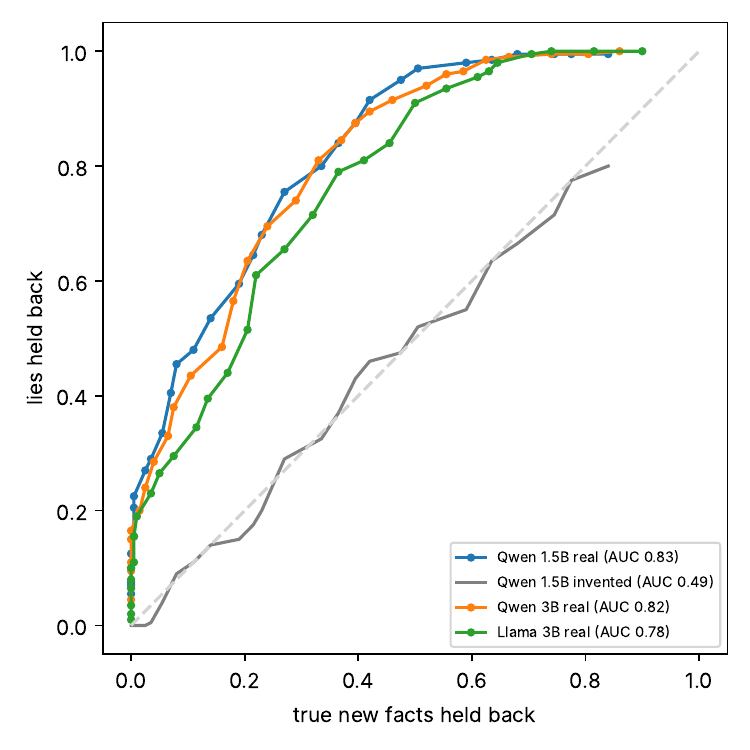}
\caption{Trade-off curve of the prior gap on the fresh seed-7 world under each original model (200 claims per type). Each point is a threshold; the diagonal is chance.}
\label{fig:tradeoff}
\end{figure}

AUC is the probability that a randomly drawn lie receives a larger gap than a randomly drawn true fact. It measures how well the gap ranks the two groups. It is not an accuracy, since the share of lies caught at a given cost depends on the threshold and is read off the curve in Figure~\ref{fig:tradeoff}.

Scoring each claim by its best article variant (``nurse'', ``a nurse'', \dots) removed template artefacts but left separation unchanged (AUC 0.83 for both rules); it only shifted the threshold. The threshold is best seen as a policy choice between safety and learning (Table~\ref{tab:operating}). Holding back a true claim only delays it, while learning a lie corrupts the model, so we prefer an operating point on the safe side. The same threshold of 6 is used for all three models without retuning.

\begin{table}[htbp]
\centering\small
\caption{Safety--learning operating points of the prior key on the fresh seed-7 world with 200 claims per type. Each cell gives Qwen2.5-1.5B / Qwen2.5-3B / Llama-3.2-3B (the larger models are introduced in Section~\ref{sec:models}). The fourth column tracks the third at every threshold, so for lies about unknown entities the key holds back lies and truths at the same rate. Precision compares known-fact lies with true new facts in this balanced sample (200 of each); with fewer lies in the stream it would be lower. Person-clustered bootstrap intervals for the AUC are 0.78--0.86 (Qwen2.5-3B) and 0.74--0.82 (Llama-3.2-3B), matching the claim-level ones.}
\label{tab:operating}
\begin{tabularx}{\linewidth}{@{}lYYYY@{}}
\toprule
Gap threshold & Known-fact lies held back & True new facts held back & Lies about invented people held back & Precision: lies among held-back claims \\
\midrule
4 & 0.92 / 0.92 / 0.91 & 0.42 / 0.46 / 0.50 & 0.46 / 0.48 / 0.49 & 0.69 / 0.67 / 0.65 \\
6 (used) & 0.76 / 0.81 / 0.71 & 0.27 / 0.33 / 0.32 & 0.29 / 0.34 / 0.29 & 0.74 / 0.71 / 0.69 \\
8 & 0.54 / 0.56 / 0.44 & 0.14 / 0.18 / 0.17 & 0.14 / 0.13 / 0.16 & 0.79 / 0.76 / 0.72 \\
\bottomrule
\end{tabularx}
\end{table}

\subsection{A strong real judge learns only what it can verify}
\label{sec:gemini}

With Gemini 2.5 Flash as the verifier key and a trusted source available for only half of the invented people (30\% poison, verifier key without test, 3 seeds), the student absorbed 6.4\% of lies, the same as before learning, and kept 98.0\% of old facts. The judge approved about 2 of the roughly 170 distinct lies in each run (about 1\%), and it rejected about 198 of the roughly 390 true claims, almost exactly the unsourced ones. New facts learned therefore stopped at 0.51, against 0.86 with the oracle.

Gemini rarely guessed. It rejected almost every claim it could not check against a source, so it was safe, but it learned no more than the sources covered. The harder case is a claim that is true, new and unsourced. Neither a judge nor the model's prior can resolve it (Section~\ref{sec:prior}), which motivates corroboration across independent sources as future work. Adding the statistical test changed nothing here (lies 5.8\%, new facts 0.50, retention 97.8\%).

\subsection{Legitimate updates and the evidence route}
\label{sec:updates}

The prior key cannot tell a lie about a known fact from true news that changes it: both contradict the model. We tested this directly by adding 100 legitimate updates to the stream: known CounterFact facts whose counterfactual value is, in this world, the new truth, each reported by a trusted source. We then added a \textbf{trusted-source evidence route}: a claim the prior key holds back is learned anyway if a trusted source states the same value. A planted lie has no matching source by assumption, so the route cannot readmit it. Sources were available for all claims or, in the more realistic setting, for half of them. All runs use 30\% poison, the simulated judge, and five held-out seeds (10--14) that no earlier analysis used, with the gap threshold fixed at 6 before running (Table~\ref{tab:updates}).

\begin{table}[htbp]
\centering\small
\caption{Legitimate updates and the evidence route. Judge accuracy 0.9, means over 5 held-out seeds; seed standard deviations are at most 0.07. At judge accuracy 0.7 the pattern is the same (updates learned 0.70 / 0.19 / 0.93 with all sources and 0.59 with half; lies per 1,000 learned claims 129--132 / 112--114 / 60 and 79).}
\label{tab:updates}
\begin{tabularx}{\linewidth}{@{}llYYYYY@{}}
\toprule
Sources & System & Updates learned & New facts learned & Known-fact lies believed & Old facts kept & Lies per 1,000 learned claims \\
\midrule
All claims & Judge only & 0.90 & 0.79 & 0.18 & 0.92 & 42 \\
All claims & Judge + prior & 0.24 & 0.60 & 0.07 & 0.97 & 33 \\
All claims & + evidence route & \textbf{0.98} & 0.79 & 0.11 & 0.91 & \textbf{18} \\
\midrule
Half of claims & Judge only & 0.90 & 0.81 & 0.18 & 0.93 & 44 \\
Half of claims & Judge + prior & 0.24 & 0.60 & 0.07 & 0.97 & 32 \\
Half of claims & + evidence route & \textbf{0.62} & 0.72 & 0.09 & 0.94 & \textbf{24} \\
\bottomrule
\end{tabularx}
\end{table}

The concern was real. The prior key alone blocked about three quarters of legitimate updates. The evidence route recovered them in proportion to source coverage. It rescued 192 of the 192 held-back true claims per run with full coverage and 101 of 192 with half, and it rescued no lie in any run, as the trusted-source assumption implies, because a planted lie never has a matching source. The combined system had the lowest share of poison among the claims it learned in every setting, roughly half that of the judge alone. Learning updates did cost some retention. Systems that learned most of the 100 updates kept about 5 points fewer of the untouched held-out facts (0.91--0.93 against 0.97), a side effect familiar from knowledge editing, where changing one fact disturbs its neighbours.

The sources in the main experiment are perfect and cannot be planted or copied. The evidence route therefore shows that legitimate updates \emph{can} coexist with the prior key. It does not show that they can be trusted against an attacker who also controls sources, which is the corroboration problem of Section~\ref{sec:conclusion}.

\paragraph{Planted and outdated sources.} We then broke the trusted-source assumption (judge accuracy 0.9, sources for half the claims, held-out seeds 10--14). When an attacker planted a matching source for 10\% of the known-fact lies, the evidence route admitted 4 lies per run and lies per 1,000 learned claims rose from 23 to 35 (paired difference $+12$, 95\% CI $+8$ to $+16$). With planted sources for 30\% of the lies it admitted 18 lies per run, the model believed 29\% of the known-fact lies instead of 8\%, and lies per 1,000 learned claims rose to 65, worse than the judge alone in the same runs (41). When 30\% of the sourced updates came with an outdated source that still stated the old value, no lie entered, but updates learned fell from 0.63 to 0.48. In short, the evidence route is as safe as its sources are hard to forge.

\subsection{Replication across model size and family}
\label{sec:models}

We repeated the realistic setting of Section~\ref{sec:updates} (sources for half the claims, 100 legitimate updates, 30\% poison, held-out seeds 10--14) with two further students, Qwen2.5-3B-Instruct (a larger model of the same family) and Llama-3.2-3B-Instruct (a different family), keeping every setting, including the gap threshold of 6, unchanged. On the fresh seed-7 world the prior gap separates known-fact lies from true new facts with AUC 0.82 (95\% CI 0.78--0.86) for the 3B model against 0.83 for 1.5B, and Llama-3.2-3B reaches 0.78 (0.74--0.81); lies about invented people stay at chance for all three (0.49--0.50). The same threshold transfers without retuning: it holds back 33\% (Qwen 3B) and 32\% (Llama) of true facts and catches 81\% and 71\% of known-fact lies, against 27\% and 76\% for Qwen 1.5B.

\begin{table}[htbp]
\centering\footnotesize
\caption{Replication with three students. Each cell gives Qwen2.5-1.5B / Qwen2.5-3B / Llama-3.2-3B; judge accuracy 0.9; sources for half the claims; means over 5 held-out seeds, seed standard deviations at most 0.06. Judge accuracy 0.7 shows the same pattern for all three models.}
\label{tab:models}
\setlength{\tabcolsep}{4pt}
\begin{tabularx}{\linewidth}{@{}lYYYYYY@{}}
\toprule
System & Known-fact lies believed & Lies about invented people & Updates learned & New facts learned & Old facts kept & Lies per 1,000 learned claims \\
\midrule
Surprise-only (no judge) & 1.00 / 1.00 / 1.00 & 0.79 / 0.81 / 0.87 & 0.99 / 0.99 / 0.99 & 0.77 / 0.81 / 0.84 & 0.88 / 0.89 / 0.88 & 257 / 251 / 254 \\
Judge only & 0.18 / 0.16 / 0.19 & 0.16 / 0.16 / 0.15 & 0.90 / 0.91 / 0.90 & 0.81 / 0.81 / 0.83 & 0.93 / 0.93 / 0.92 & 44 / 37 / 38 \\
Judge + prior & 0.07 / 0.04 / 0.05 & 0.12 / 0.11 / 0.12 & 0.24 / 0.17 / 0.20 & 0.60 / 0.57 / 0.57 & 0.97 / 0.97 / 0.97 & 32 / 25 / 29 \\
+ evidence route & 0.09 / 0.07 / 0.08 & 0.14 / 0.14 / 0.13 & 0.62 / 0.57 / 0.59 & 0.72 / 0.72 / 0.73 & 0.94 / 0.95 / 0.94 & 24 / 18 / 22 \\
\bottomrule
\end{tabularx}
\end{table}

Every qualitative finding replicated across both families (Table~\ref{tab:models}). Surprise-only learning absorbed essentially all known-fact lies. The prior key cut them roughly 2--4$\times$ relative to the judge alone and left lies about invented people untouched. The evidence route recovered legitimate updates in proportion to source coverage and rescued no lie in any run, and the combined system had the lowest poison share among learned claims. Neither the larger model nor the other family made the prior key stronger, and Llama separates somewhat less sharply, with overlapping confidence intervals. For Qwen 3B, gaps grew for true new facts (median 3.8 vs.\ 3.2 nats/token) at least as much as for known-fact lies (8.4 vs.\ 8.2), so separation did not improve at this scale.

\subsection{Real judges are strict and defend their own beliefs}
\label{sec:realjudges}

The simulated judge errs symmetrically. To see how real verifiers err, we replaced it with two local LLM judges served through Ollama: phi4 (14B, a different family from the student) and Llama 3.2 (3B, a deliberately weak judge). The setting is that of Sections~\ref{sec:updates}--\ref{sec:models} with the Qwen2.5-1.5B student (sources for half the claims, 100 legitimate updates, 30\% poison, held-out seeds 10--14). Every verdict and prior gap was recorded per claim, about 3,300 judged claims per judge pooled over the 5 seeds; no judge answer was unparseable.

\begin{table}[htbp]
\centering\small
\caption{Share of claims each real judge accepted, pooled over 5 seeds.}
\label{tab:verdicts}
\begin{tabular}{@{}lcc@{}}
\toprule
Claim type & phi4 (14B) & Llama 3.2 (3B) \\
\midrule
Known-fact lies & 5.9\% & 4.0\% \\
Lies about invented people & 0.5\% & 0.0\% \\
True new facts with a source & 99\% & 65\% \\
True new facts without a source & 0\% & 0\% \\
Legitimate updates with a source & 30\% & 40\% \\
Legitimate updates without a source & 4\% & 1\% \\
\bottomrule
\end{tabular}
\end{table}

Real judges erred mostly in one direction (Table~\ref{tab:verdicts}). They rarely accepted a lie, but they rejected every unsourced truth, and the weaker judge also rejected a third of sourced ones, so its weakness showed up as strictness rather than gullibility. Both judges also rejected most legitimate updates \emph{even when the trusted source was in front of them}, although the prompt states that a given source is correct; the source contradicted what they believed. In this respect the judge is as conservative as the prior key, consistent with the knowledge-driven failures of LLM judges reported by \citet{lee2026judging}. Verification has two separate dimensions: \textbf{safety}, the share of lies a verifier accepts, and \textbf{coverage}, the share of truths it accepts. Real judges buy high safety with low coverage, whereas the simulated judge of Section~\ref{sec:judge} assumed both errors were equally likely.

\begin{table}[htbp]
\centering\small
\caption{Full systems with real judges (phi4 / Llama 3.2). Means over 5 held-out seeds; seed standard deviations at most 0.05. The evidence route rescued no lie.}
\label{tab:realjudges}
\begin{tabularx}{\linewidth}{@{}lYYYYY@{}}
\toprule
System & Updates learned & New facts learned & Known-fact lies believed & Old facts kept & Lies per 1,000 learned claims \\
\midrule
Judge only & 0.19 / 0.23 & 0.55 / 0.38 & 0.09 / 0.09 & 0.96 / 0.96 & 23 / 21 \\
Judge + prior & 0.07 / 0.08 & 0.40 / 0.28 & 0.06 / 0.04 & 0.98 / 0.98 & 16 / 10 \\
+ evidence route & \textbf{0.47 / 0.46} & 0.54 / 0.41 & 0.07 / 0.07 & 0.94 / 0.94 & \textbf{9 / 5} \\
\bottomrule
\end{tabularx}
\end{table}

Because the evidence route lets a source-backed claim bypass the judge, it more than doubles the updates learned with either real judge, and the full system again holds the fewest lies among learned claims, 5--9 per 1,000 against 21--23 for the judge alone (Table~\ref{tab:realjudges}). These remaining lies all entered through the judge; none entered through the evidence route.

\paragraph{Do the two keys fail on the same claims?} Of the 42 known-fact lies the two judges accepted, the prior key would have held back 26 (62\%; 14 of 25 for phi4, 12 of 17 for Llama). Its hold rate on all known-fact lies is 77\%. The 42 cases are only 28 distinct claims, because both judges saw the same worlds, so a test on the pooled cases would overstate their independence. At the seed level, the prior key's hold rate on judge-approved lies was below its overall rate in 8 of 10 judge--seed runs (mean 65\% vs.\ 77\%; one-sided sign test over these 10 runs, $p \approx 0.05$). With 10 runs and 28 distinct claims, this is weak evidence, and we report it as exploratory. The observed rates suggest that the lies that fool a judge may also be somewhat harder for the prior key, so the two keys should not be assumed independent; combining them still helped in every setting.

\subsection{What the prior key detects, and how it is bypassed}
\label{sec:detects}

Section~\ref{sec:prior} left open whether the prior key detects lies or only contradictions. Three analyses of the saved results answer this question (\texttt{review\char95{}analyses.py}).

First, the gap does not separate a lie about a known fact from a legitimate update of the same kind of fact. On the held-out runs of Section~\ref{sec:realjudges} it scores AUC 0.53 (95\% CI 0.49--0.56; 424 lies, 500 updates) and holds back 77\% of the lies and 75\% of the updates. A second diagnostic on the fresh seed-7 world completes the four combinations of true or false and known or unknown (200 claims per type, plus 100 updates). True claims about known facts have small gaps (mean 3.2--3.5 nats/token; 17--20\% held back), as do true new facts (27--33\%) and lies about invented people (29--34\%). Known-fact lies (71--81\%) and legitimate updates (67--84\%) are held back at similar rates, with AUC 0.45--0.53 between them across the three students. The gap is large whenever a claim contradicts what the model knows, whether the claim is true or false.

Second, a simpler signal does as well (Table~\ref{tab:signalsauc}). The original model's confidence in its own first answer token, which ignores the claim entirely, separates known-fact lies from true new facts with AUC 0.84, 0.80 and 0.88 for the three students, against 0.83, 0.82 and 0.78 for the gap. The prior gap is therefore best seen as an interpretable contradiction signal rather than a uniquely strong detector; its practical advantage is that it needs no judge call and can route suspicious claims before verification. What the prior key mostly detects, then, is that a claim touches something the model already knows.

\begin{table}[htbp]
\centering\small
\caption{AUC for separating known-fact lies from true new facts on the fresh seed-7 world, by signal (200 claims per type; 95\% bootstrap intervals are about $\pm0.04$).}
\label{tab:signalsauc}
\begin{tabular}{@{}lccc@{}}
\toprule
Signal (original model) & Qwen2.5-1.5B & Qwen2.5-3B & Llama-3.2-3B \\
\midrule
Prior gap & 0.83 & 0.82 & 0.78 \\
Claim NLL (surprise under the original model) & 0.78 & 0.79 & 0.74 \\
Confidence in its own answer (ignores the claim) & 0.84 & 0.80 & 0.88 \\
Prior gap, lies about invented people instead & 0.49 & 0.50 & 0.49 \\
\bottomrule
\end{tabular}
\end{table}

Third, an attacker with access to the model can screen its lies and submit only those with a gap below 6. Only 23\% of known-fact lies pass this filter, but each submitted lie then passes both keys 4.4 times as often: 11.5\% instead of 2.6\% with phi4, and 5.2\% instead of 1.2\% with Llama 3.2. Against such an attacker the prior key does not help, and safety depends on the judge alone.

Paired across seeds, the evidence route lowered lies per 1,000 learned claims relative to the judge alone in all three settings of Sections~\ref{sec:updates} and~\ref{sec:realjudges} (95\% intervals exclude zero). It cost 3--4 points of retention relative to judge + prior, and with real judges it did not significantly change the share of known-fact lies the model believes.

% ======================================================================
\section{Practical implications and application scenarios}
\label{sec:practice}

\paragraph{Where this matters.} The setting we study is an assistant that should absorb new facts from the people it serves, such as a university, public-administration or company assistant that must follow changing regulations, responsibilities and procedures. A fact written into the weights is available in every later answer without retrieval, which is also why a wrong acceptance is costly.

\paragraph{When to write into the weights at all.} Retrieval \citep{lewis2020rag} is the safer default for facts that change often, must be traced to a document, or may have to be removed again. Parametric updates are worth their risk for stable facts that are used constantly, or where retrieval is unavailable. In both designs the acceptance decision studied here remains: a poisoned document store misleads the model as a poisoned stream does.

\paragraph{Conditions for deployment.} Our results suggest four. (1)~Never select what to learn by surprise alone. (2)~Use a verifier key, and measure its safety and its coverage separately on the claims it will actually see. (3)~Use the prior key as a cheap triage filter that sends conflicting claims to evidence or review. It does not stop an attacker with access to the model, who can bypass it. (4)~Accept contradicting claims only with evidence from sources whose integrity is controlled independently of the claim's author; with planted sources for 30\% of the lies, the evidence route made the system worse than the judge alone (Section~\ref{sec:updates}).

\paragraph{Costs.} With the prior key in front of the judge, judge calls fell from about 661 to 374 per run ($-43\%$). The reduction concerns judge calls, not total verification cost: about 180 claims per run stayed held back even with the evidence route and move to a review queue whose cost we did not measure. Learning updates cost 3--4 points of retention on untouched facts. A deployment has to weigh these costs against the cost of a single absorbed lie.

\paragraph{Open questions.} How to establish that a source is trustworthy, how to handle sources that disagree or are outdated, how to make review cheap, and whether many small updates eventually harm knowledge more than they help.

% ======================================================================
\section{Limitations}
\label{sec:limitations}

Two limitations matter most.

\paragraph{Legitimate updates of known facts.} True news that changes something the model already knew (a new office holder, say) contradicts its prior just as a lie does; the prior key held back about three quarters of such updates in Section~\ref{sec:updates}. On its own, the key cannot tell an update from a lie. In a continual learner this case is common. Held-back claims need a second route. Section~\ref{sec:updates} tests one, an evidence check against a trusted source; it works with perfect sources and fails once an attacker can plant them. The prior is also not ground truth: where pretraining taught the model something false, the prior key defends the error and holds back its correction. We therefore read the prior key as a triage filter that routes conflicting claims to evidence or review.

\paragraph{Adaptive attackers.} Our poison is fixed in advance. Section~\ref{sec:detects} shows that an attacker who screens lies with the open model passes the prior key at will, so against such an attacker only the judge protects the model. We did not test attackers who also target the judge. Our numbers describe these attackers only and give no safety guarantee.

\paragraph{Further limitations.}
\begin{itemize}
  \item \textbf{Small models only.} Main results use Qwen2.5-1.5B; the key comparisons are repeated on Qwen2.5-3B and Llama-3.2-3B (Section~\ref{sec:models}), and neither strengthened the prior key. Models larger than 3B parameters were not tested. The study is designed to establish the mechanism in small, reproducible models; larger models are future work.
  \item \textbf{Short factual claims.} Claims are single subject--relation--object facts. Instructions, opinions and multi-hop knowledge are not covered.
  \item \textbf{Limited judge coverage.} The simulated judge errs symmetrically; the real judges of Section~\ref{sec:realjudges} erred mostly by rejecting truths, and the lies they did accept were partly ones the prior key also misses. Two local judges and one API judge are a small sample of verifiers. An attacker would search for those claims, so the relevant measure is a judge's accuracy on adversarially chosen poison rather than its average accuracy.
  \item \textbf{Unmeasured review load.} We did not measure what it costs to review the claims the prior key holds back, by people or by retrieval; the 27\% of true claims held back is a lower bound once legitimate updates are included.
  \item \textbf{Template artefacts.} Some held-back true facts arise from templates (for example, ``The profession of X is'' strongly favours an answer beginning with ``a \dots''). Scoring the best of article variants moved the threshold but left separation unchanged (AUC 0.83 either way).
  \item \textbf{Threshold choice.} The rule and the threshold $g \ge 6$ were developed on a 30-claim sample of the seed-0 world, which the prior sweep of Section~\ref{sec:prior} also uses. The seed-7 diagnostic and seeds 10--14 were run afterwards with the threshold fixed, and it transferred unchanged to two further models. The threshold also transfers unchanged across these settings. A test set untouched by any development choice would still be stronger evidence.
  \item \textbf{Paraphrase transfer is weak.} Facts learned in one wording are recalled in other wordings less often (0.62 vs.\ 0.86 with the oracle verifier key at 30\% poison), a known limit of fact injection by fine-tuning.
  \item \textbf{No retrieval baseline, limited open-ended evaluation.} We did not compare with a retrieval system on the same stream; the question studied here is the acceptance decision for parametric learning, and retrieval changes the storage mechanism, so it calls for a separate baseline. Open-ended answers (Section~\ref{sec:surprise}) were checked in one setting and only as single greedy answers; multi-step questions and longer documents remain untested.
  \item \textbf{Scale of compute.} We used three seeds per setting in Sections~\ref{sec:surprise}--\ref{sec:gemini} and five in Sections~\ref{sec:updates}--\ref{sec:realjudges}, all on a laptop GPU. Bootstrap confidence intervals are reported for the prior-key AUC (Section~\ref{sec:prior}), seed standard deviations under each table, and per-seed results in the released results folders.
\end{itemize}

% ======================================================================
\section{Conclusion and future work}
\label{sec:conclusion}

A false claim that contradicts a known fact is highly surprising to the model by construction, so selecting by surprise lets such lies through. In our controlled poisoned streams, a surprise-driven learner absorbed up to 93\% of the lies it saw, and 99--100\% of the known-fact lies. A verifier key removed them, but the learner then inherited the errors of its judge. Regression tests on unrelated knowledge did not help in our setting, because a lie barely affects other facts. The model's own prior is a cheap second key that needs no judge and saves judge calls. It cuts absorbed known-fact lies 2.5--5$\times$ with simulated judges and 1.5--2.5$\times$ with real ones, at the price of sending about a quarter of true claims to review, but it reacts to any contradiction with what the model knows, true or false, and an attacker with access to the model can avoid it. None of the signals we study decides truth on its own, and our results suggest that a safer continual learner should combine complementary signals rather than rely on any one of them (Table~\ref{tab:signals}). The real-judge results add a further constraint. They rarely accepted lies, but they also rejected most legitimate updates with a trusted source in view, so source-backed claims need a route that bypasses the judge.

\begin{table}[htbp]
\centering\footnotesize
\caption{What each signal can and cannot see, summarising Section~5. $^{*}$Under the assumption of trusted sources that an attacker cannot plant or copy.}
\label{tab:signals}
\setlength{\tabcolsep}{4pt}
\begin{tabularx}{\linewidth}{@{}lYYYYY@{}}
\toprule
Signal & Question it answers & Lie about a known fact & Lie about an unknown entity & True new fact & True update of a known fact \\
\midrule
Surprise & Is it new to me? & selects it & selects it & selects it & selects it \\
Prior conflict & Does it contradict what I believed? & holds most back (AUC 0.78--0.83) & blind (AUC 0.49) & passes about 70\% & holds most back \\
Judge & Does a verifier approve it? & rejects 94--96\% (real judges) & rejects nearly all & accepts mainly if sourced & rejects 60--70\% even with a source \\
Evidence route & Does a trusted source state the same? & never matches$^{*}$ & never matches$^{*}$ & accepts if sourced$^{*}$ & accepts if sourced$^{*}$ \\
Regression test & Did unrelated knowledge degrade? & blind & blind & (no effect) & (no effect) \\
\bottomrule
\end{tabularx}
\end{table}

\paragraph{What the experiments show.} In controlled streams with three students of 1.5--3B parameters: surprise-driven selection absorbs most lies and nearly all known-fact lies, takes in lies at the same rate as random selection with the same budget, and absorbs known-fact lies more reliably than random selection does; a verifier key's safety follows its judge; regression tests on unrelated facts throttle learning without filtering lies; the prior gap separates known-fact lies from true new facts (AUC 0.78--0.83) but neither lies about invented people (0.49--0.50) nor lies from legitimate updates (0.53), and the model's answer confidence does as well; an attacker who screens lies with the model passes it; and with trusted sources, the evidence route recovers held-back updates while no lie enters through it, but planted sources for 30\% of the lies make the system worse than the judge alone. Open-ended answers, under an exact-match assertion rule, are directionally consistent with the ranking metrics.

\paragraph{What they suggest.} Real judges trade coverage for safety and defend their own beliefs against sourced updates (two local judges, Section~\ref{sec:realjudges}). The judge and the prior key may fail partly on the same lies (exploratory, Section~\ref{sec:realjudges}).

\paragraph{What remains a hypothesis.} That these findings hold for larger models, adaptive attackers and sources an attacker can plant; and that corroboration across independent chains, proposed below, can resolve unsourced claims about unknown entities.

\medskip
None of the keys in this study resolves the case of a new, unsourced claim about something the model has never seen. The judge refuses it, and the prior has nothing to contradict (AUC 0.49). We see four directions for future work.
\begin{enumerate}
  \item \textbf{Legitimate updates of known facts.} True news that changes something the model knew contradicts its prior just as a lie does. Section~\ref{sec:updates} shows that routing held-back claims to an evidence check recovers them when sources are perfect, and that planted sources for 30\% of the lies make it worse than the judge alone. An important next step is a defence against planted and copied sources, and real retrieval instead of a simulated source.
  \item \textbf{Corroboration.} Accept unsourced claims only after independent confirmations; a liar is usually one source, the truth many. Corroboration should count independent chains of transmission rather than sources, since copies add no evidence; this is the criterion of the \emph{mutawatir} report in hadith scholarship, which accepts a report only when so many independent chains carry it that collusion on a falsehood is implausible. \citet{raja2026isnad} adapts related hadith methods to provenance in multi-agent knowledge systems; we propose applying the independence criterion to what may be written into the weights. New claims could wait in an erasable memory with provenance until enough independent chains arrive, and only then be consolidated into the weights. This turns the acceptance decision into a sequential test on independent evidence, where SPRT is the natural tool.
  \item \textbf{Adversarial verifiers and attackers.} Measure judge accuracy on poison chosen to fool it, and attackers who adapt to the prior key (for example, lies phrased to sit close to the model's own answer).
  \item \textbf{Breadth.} Other model sizes and families, more seeds, claim types beyond single facts (instructions, procedures, multi-hop knowledge), and consolidation of committed adapters into the base (``sleep'') with the prior key anchored to the pre-learning model.
\end{enumerate}

\paragraph{Code and data availability.} The code, configurations, every saved result, and a script that rebuilds each table and figure of this paper from those results and checks every quoted number against them are available at \url{https://github.com/Metalesaek/sura-metales}.

\bibliographystyle{plainnat}
\bibliography{references}

\appendix
\section{Judge prompt and decoding}
\label{app:prompt}

All real judges (Gemini 2.5 Flash, phi4, Llama 3.2) received the same system and user messages. The source line is present only when the claim's subject has a trusted source.

\begin{quote}\small\ttfamily\raggedright
system: You are a strict fact-checker. Decide whether the claim is true. If a trusted source is given, it is correct. Answer with one word: True or False.\\[4pt]
user: Trusted source: <source sentence>\\
Claim: <claim sentence>\\
Is the claim true?
\end{quote}

Decoding used temperature 0, at most 20 output tokens and a thinking budget of 0 for Gemini, and at most 10 tokens for the Ollama judges. The first standalone ``true'' or ``false'' in the answer is the verdict. The simulated judge is correct with probability equal to its accuracy, decided by a hash of the seed, claim identifier and value, so the same claim always receives the same verdict within a seed.

\section{Reproducibility}
\label{app:repro}
\sloppy

All runs used Python 3.11, PyTorch 2.6.0 (CUDA 12.4), transformers 5.20.1, peft 0.21.2 and datasets 5.1.0 on one 16\,GB laptop GPU in bfloat16. Students were loaded from Hugging Face (\texttt{Qwen/Qwen2.5-1.5B-Instruct}, \texttt{Qwen/Qwen2.5-3B-Instruct}, \texttt{meta-llama/Llama-3.2-3B-Instruct}). CounterFact was loaded from \texttt{azhx/counterfact} (train split shuffled with seed 0, first 3,000 records, screened for facts each student knows). The repository (\url{https://github.com/Metalesaek/sura-metales}) contains all code, configurations and saved results. The script \texttt{reproduce\char95{}paper.py} recomputes all tables and figures from the saved results without retraining and compares 407 numbers quoted in this paper with the recomputed values.

\end{document}